\documentclass[11pt,a4paper]{article}
\usepackage[hyperref]{acl2018}
\usepackage{times}
\usepackage{latexsym}
\usepackage{amsmath}

\usepackage{url}

\aclfinalcopy 

\title{Imbalanced multi-label classification using multi-task learning with extractive summarization}

\author{John Brandt \\
  Yale University / New Haven, CT 06511 \\
  {\tt john.brandt@yale.edu} \\}

\date{}

\begin{document}
\maketitle
\begin{abstract}
  Extractive summarization and imbalanced multi-label classification often require vast amounts of training data to avoid overfitting. In situations where training data is expensive to generate, leveraging information between tasks is an attractive approach to increasing the amount of available information. This paper employs multi-task training of an extractive summarizer and an RNN-based classifier to improve summarization and classification accuracy by 50\% and 75\%, respectively, relative to RNN baselines. We hypothesize that concatenating sentence encodings based on document and class context increases generalizability for highly variable corpuses.\footnote{Code and data is available on the author's \href{https://github.com/JohnMBrandt/text-classification}{GitHub}.}
\end{abstract}

\section{Introduction}

Multi-label sentence classification is a common applied task in natural language processing. Downstream applications include the identification of themes and topics in documents, quantifying differences between documents, emotion detection, and document labelling. In this paper, we seek to address the application of identifying and tagging relevant sentences across many documents of the same domain. We build upon a recurrent neural network classifier by leveraging context information from the source documents by jointly training a surrogate extractive summarizer. It is hypothesized that concatenating sentence encoding with context information encoded by the summarization hidden layers parallels the natural decision making process undertaken by human classification, which uses some combination of semantics and context.

\section{Methods}

\subsection{Model Architecture}

The model architecture is comprised of an extractive summarizer and a sentence classifier (Figure 1). In the extractive summarizer, pre-trained GloVe embeddings \cite{pennington2014glove} are concatenated with a learned embedding layer and are fed sequentially to a bi-directional gated recurrent unit (GRU) layer \cite{cho-al-emnlp14}. The bidirectional layer contains both a forward GRU, reading words sequentially from $w_1$ to $w_n$, and a reverse GRU, reading words sequentially in reverse from $w_n$ to $w_1$. Attention is applied over the words to encode the sentences using the equations in (1-3) proposed by Yang et al.  \shortcite{Yang2016HierarchicalAN}. 

\begin{align}
u_{it} &= \text{tanh}(W_w h_{it}+b_s), \\
\alpha_i &= \frac{\text{exp}(u_{it}^Tu_w)}{\sum_t \text{exp}(u_{it}^T u_w}, \\
s_i &= \sum_t \alpha_{it}h_{it}
\end{align}

where $W$, $h$, and $b$ refer to the weight matrix, hidden state, and bias, respectively, and $\alpha$ is a learned parameter weighing each hidden state. Then, a sentence-level GRU incorporates cross-sentence context in a binary classification setting to identify whether the sentence is relevant. Here, relevance is determined as whether the sentence belongs to any of the classes. 

The sentence classifier passes pre-trained GloVe embeddings to a bi-directional GRU layer. After applying attention over the hidden states, a sigmoid activation is used to classify multiple labels for each sentence. The encoding output from the sentence GRU in the document summarizer is shared with the output of the word-level GRU in the sentence classifier. This is done by passing a matrix  identifying, for each sample, the position of the relevant encoding in the summarizer output matrix. A lambda layer is used to slice the summarizer output and share the summarizer sentence representation with the classifier network as shown in (4), where $h_e$ is the extractor GRU output, $h_c$ is the classifier GRU output, and $i$ is the index of the $h_e$ matrix corresponding to $h_c$.

\begin{equation}
(i h_{e[:,i,:]}) \oplus h_c
\end{equation}

\begin{figure*}[!ht]
\includegraphics[width= 6in]{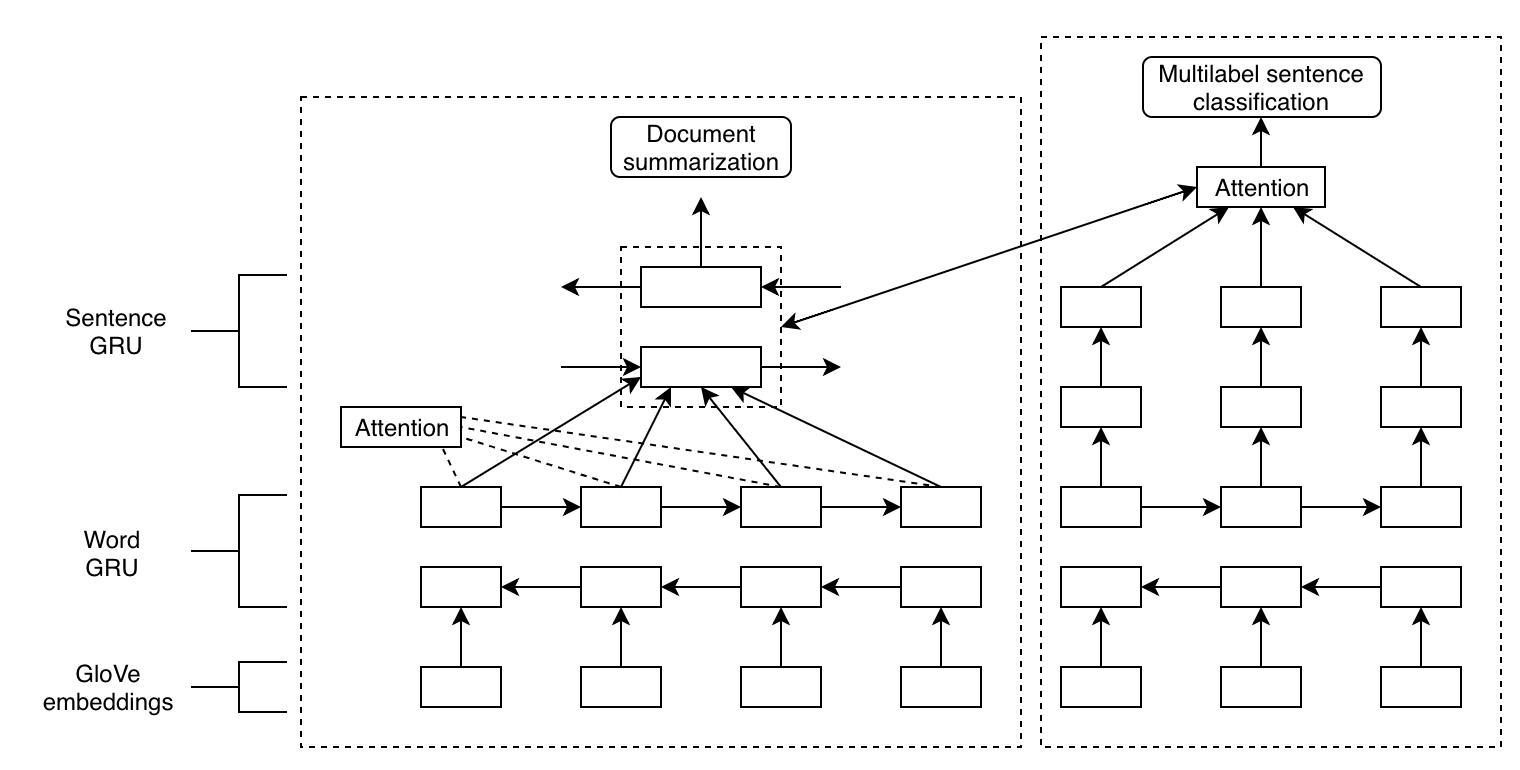}
\caption{Model architecture comprised of document summarization (left) and multi-label sentence classification (right) indicating the shared layer and the position of attention layers.}\label{fig:quintile}
\end{figure*}

\subsection{Training}

Sentences were tokenized with a max word count of 10,000 and encoded with pre-trained GloVe embeddings. Contractions and punctuation were treated as their own word. Sentences were padded to 30 words. Training was done on a Linux cluster with TensorFlow, using the Adam optimizer with a learning rate of $2.5e-5$. The sentence extractor GRU was regularized with an l1-l2 regularizer ($1e-5$). Classes were weighted by their inverse frequency for both the extractor and the summarizer inputs. In order to maintain equal sample sizes for each input, documents were broadcasted to share dimensions with classified sentences. For each document and epoch, the extractor weights were thus updated $n$ times, where $n$ is the number of relevant sentences in the document.  The loss function is the sum of the binary cross entropy loss for each output (5).

\begin{equation}
- \sum y_s(x) \log y_s(x) - \sum y_d \log y_d(x)
\end{equation}

The architecture was initially built with a 60/20/20 training, validation, test split. All results presented in this paper use an 80/20 training test split.

\subsection{Data}
\label{sect:pdf}

\begin{table}[t!]
\begin{center}
\begin{tabular}{l l}
\hline \bf Class & \bf Samples \\ \hline 
No Poverty & 15  \\
Zero Hunger & 850 \\
Good Health &  1042 \\
Education & 322 \\
Gender Equality & 367 \\
Clean Water & 841 \\
Clean Energy & 2258 \\
Economic Growth & 241 \\
Infrastructure & 579 \\
Reduced Inequality & 65 \\
Sustainable Cities & 591 \\
Responsible Production & 625 \\
Climate Action & 1106 \\
Aquatic Life & 283 \\
Land Life &  1056 \\
Peace and Justice & 121 \\
Partnerships &  406 \\
\hline
\end{tabular}
\end{center}
\caption{\label{font-table} Number of samples per class. Note that the sum of the samples exceeds the sample size due to the multi-label nature of the data.}
\end{table}

The data set was aggregated by Resource Watch \cite{RW}. Sentence relevance to 17 separate classes were hand-coded for 151 national environmental policies, totalling approximately 4,000 pages, by a team of domain experts. More specifically, sentences and phrases were classified according to their relevance to the Sustainable Development Goals, a set of 17 goals adopted by the United Nations to guide global development agenda. Class labels are shown in Table 1. There is significant overlap between the classes, and many sentences and phrases are identified as belonging to multiple classes.

The original data set only included the extracted sentences and phrases, as well as the raw HTML files of the policies. Extracted phrases were matched to their location in source documents using a combination of regular expression and the R package stringR. Four of the the 155 original documents were not considered because of differences between source and extracted sentences. All of the 151 documents used were matched at 100\% accuracy.

\section{Experiments}

\begin{table}[t!]
\begin{center}
\begin{tabular}{l c c c}
\hline \bf Model Type & \bf Loss & \bf Top-1 & \bf Top-3\\ \hline
GRU & 0.2626 & 0.368 & 0.601 \\
Multi-task & \bf 0.2065 & \bf 0.553 & \bf 0.791 \\
\hline
\end{tabular}
\end{center}
\caption{\label{font-table} Classification metrics for baseline (GRU) and multi-task network. }
\end{table}

\begin{table}[t!]
\begin{center}
\begin{tabular}{l c c c}
\hline \bf Model Type & \bf Loss & \bf Precision & \bf Recall\\ \hline
Hierarchical GRU & 0.400 & 0.236 & 0.312 \\
Multi-task & \bf 0.181 & \bf 0.460 & \bf 0.459 \\
\hline
\end{tabular}
\end{center}
\caption{\label{font-table} Summarization metrics for baseline (GRU) and multi-task network. }
\end{table}

\begin{figure}[!ht]
\includegraphics[width= 3in]{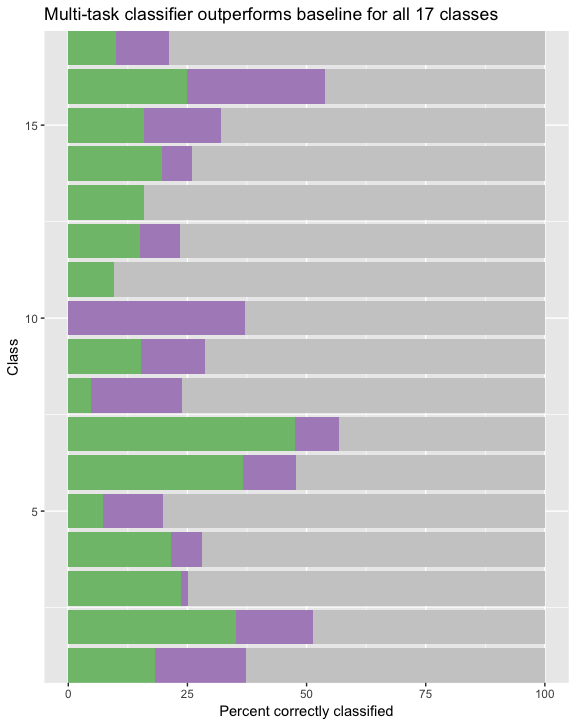}
\caption{Baseline GRU (green) and multi-task (purple) sentence classification accuracy by class demonstrates that, for each of the classes, leveraging the source document context improves accuracy.}\label{fig:quintile}
\end{figure}

To test the effectiveness of sharing weights between a sentence classifier and an extractive summarization network, an ablation study was used to test the multi-task network against its individual components. Sentence classification was tested using a GRU with attention, and extractive summarization was tested with a hierarchical GRU with attention over word embeddings. 

Classification performance was evaluated by considering loss, accuracy at top-K, and accuracy for each class, while summarization performance was evaluated with precision and recall. All models were trained with the same training, validation, and test data, and the epoch with the lowest validation loss was used for comparison. Data was split to ensure that no documents or sentences were shared between validation and test data.

The multi-task model outperforms the RNN baseline in terms of top-1 accuracy, percent accuracy by class, and predicting the number of labels for a candidate sentence. As shown in Table 2, the multi-task model increases top-1 accuracy by 20\% absolute over the RNN baseline. Figure 2 shows the accuracy by class, demonstrating that the multi-task model significantly improves upon the baseline for identifying extremely unbalanced classes. 

For predicting the number of labels, the multi-task model improves upon the RNN baseline significantly. While the RNN baseline had a false negative rate of 23\%, the multi-task model reduces this error to 13\%. As the confusion matrix in Figure 3 shows, the multi-task model also identifies the correct number of labels for 1 and 2 label sentences on average 40\% more frequently than the baseline. Both the RNN baseline and the multi-task model perform poorly when identifying 3, 4, and 5 label sentences, though these represent less than 1\% of the dataset cumulatively. 

Updating weights for extractive summarization based upon their contribution to sentence classification accuracy greatly improved the summarization performance over the hierarchical GRU baseline. As Table 3 shows, the multi-task model achieved 100\% and 48\% relative gains in precision and recall over the baseline. 

\begin{figure}[!ht]
\includegraphics[width= 3in]{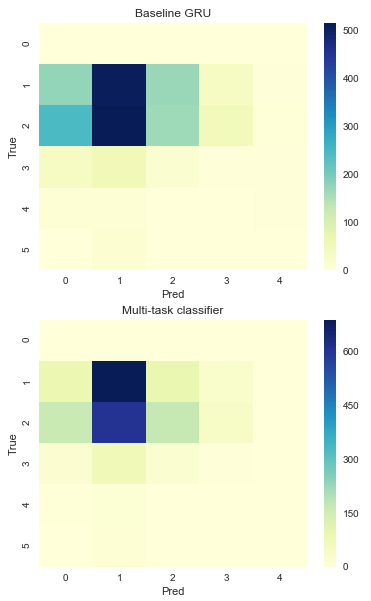}
\caption{Confusion matrix of number of topics identified by baseline and multi-task classifier indicates that the multi-task classifier predicts label number more accurately.}\label{fig:quintile}
\end{figure}

\section{Discussion}

The results suggest that, in applications where sentences are extracted from longer documents, jointly training an extractive summarization network effectively allows the classifier to leverage context information about the source documents to improve model accuracy. This is especially important for highly context-specific applications such as the present one, where the phrasing of policy agenda is dependent upon the national infrastructure, political context, and development goals, as well as the translation quality between the source language and English. 

The inclusion of an extractive summarizer allows the sentence classifier to access information about what makes a sentence relevant enough to be classified. This is a much different view of the sentences than that obtained by the sentence classifier, which simply seeks differences between classes rather than the underlying information that makes the sentence relevant. To the authors knowledge, the dataset created and presented in this paper is the largest data set containing both multi-class labels and their context in source documents. As the automated analysis of large, unstructured texts such as legal documents becomes mainstream, the results of this paper show that developing multi-task training settings where classifiers can leverage information about the source documents can improve accuracy in sparse, multi-class settings.

\bibliography{acl2018}
\bibliographystyle{acl_natbib}

\appendix

\end{document}